# MKA: A Scalable Medical Knowledge Assisted Mechanism for Generative Models on Medical Conversation Tasks

**Ke Liang**[1*]**, Sifan Wu**[2]**, Jiayi Gu**[3]

[1] Pennsylvania State University, PA 16801, USA

[2] Nvidia, Shanghai, 201210, China

[3] TMiRob, Shanghai, 201203, China

[*]Corresponding author's e-mail: kul660@psu.edu

**Abstract.** Using natural language processing (NLP) technologies to develop medical chatbots makes the diagnosis of the patient more convenient and efficient, which is a typical application in healthcare AI. Because of its importance, lots of research have been come out. Recently, the neural generative models have shown their impressive ability as the core of chatbot, while it cannot scale well when directly applied to medical conversation due to the lack of medical-specific knowledge. To address the limitation, a scalable Medical Knowledge Assisted mechanism, *MKA*, is proposed in this paper. The mechanism aims to assist general neural generative models to achieve better performance on the medical conversation task. The medical-specific knowledge graph is designed within the mechanism, which contains 6 types of medical-related information, including department, drug, check, symptom, disease, food. Besides, the specific token concatenation policy is defined to effectively inject medical information into the input data. Evaluation of our method is carried out on two typical medical datasets, MedDG and MedDialog-CN. The evaluation results demonstrate that models combined with our mechanism outperform original methods in multiple automatic evaluation metrics. Besides, MKA-Bert-GPT achieves state-of-the-art performance. The open-sourced codes are public[1].

## 1. Introduction
Difficulty in seeing a doctor, long queuing time, and inconvenience of making appointments have long been hurdles facing patients when they try to access primary care services. To solve these challenges, many advanced artificial intelligence (*AI*) technologies [1][2][3] have been combined with healthcare to boost the availability of medical resources, such as applying pattern recognition methods on medical images [4][5] and leveraging natural language processing (*NLP)* technologies to design medical chatbots [6][7]. The medical chatbot is mainly aimed to offer the medical assistants including dentification of their disease and medical suggestions for drugs, foods, checks based on their self-reports, medical front desk service that direct the patient to suitable healthcare service department, etc [8][9]. It has a significant potential to simplify the diagnostic process and relieve the cost of collecting information from patients. Besides, the preliminary diagnosis results generated by the model may assist doctors to make a diagnosis more efficiently.

---

[1] https://github.com/LIANGKE23/Knowledge_Assisted_Medical_Dialogue_Generation_Mechanism

As the core of the medical chatbot, different methods have been investigated recently. In general, typical methods can be divided into two types [10], including information retrieval-based methods and neural generative methods. As for the first type, the methods usually match the response from the user-built question and answer ($Q\&A$) pool based on the dialogue context, which means it can only provide the response occurred in the existed pool. In another word, the poor-quality pool will influence a lot on the response. The second type methods usually take the dialogue context history as input and generate the suitable response word by word. Compared to retrieval-based methods, neural generative methods are more intelligent and flexible, which is what we focus on in this paper.

Currently, different neural generative models are applied to medical domain, including *LSTM-based* models, *Transformer*, *GPT*, *BERT-GPT* and etc. However, none of them performs well on the medical domain, which is reasonable. Here is the fact that the doctor make diagnosis not only based on their experiences, but also the medical knowledge learned from professional books, especially when they meet rarely seen symptoms or diseases. The training procedure of the models only imitate the learning procedure of the experiences but leave out the learning procedure from books. However, few work is about how to effectively integrate the medical knowledge with the neural generative models. Besides, patients are usually asked to fill in the patient self-report before the conversation starts with the doctor in real-world scenario. There are two common questions in the patient self-report, including "which department do you want to go?" and "what kind of the disease or symptom do you have?". Previous medical neural generative models will either leave out the information or roughly concatenate the original context in the patient self-report with the conversation history. It may cause either information to lose or redundance problem for the methods.

To address the limitations, the objective of the paper is to propose a medical knowledge assisted mechanism, *MKA,* to assist common neural generative models to achieve better performance for the medical conversation task. *MKA* is an effective and light-weight method to integrate the medical knowledge with neural generative models. The mechanism first introduces a medical knowledge generation module to generate the related medical knowledge, which generates the medical knowledge subgraph ($MKG_{sub}$) generated from the patients' self-report. The designed knowledge graphs contain related medical knowledge for each patient, including 6 type entities (i.e., $\epsilon_{department}$, $\epsilon_{disease}$, $\epsilon_{symptom}$, $\epsilon_{food}$, $\epsilon_{check}$, and $\epsilon_{drug}$) and 6 type relations (i.e., $\gamma_{has-disease}$, $\gamma_{has-symptom}$, $\gamma_{need-drug}$, $\gamma_{need-check}$, $\gamma_{need-food}$, and $\gamma_{no-food}$). Then, the medical knowledge information is fed into the token processor together with the dialogue contexts. Within the token processor, all the tokens will be reorganized based on the specific token concatenation policy. Finally, the processed data will be taken by selected generative models for training. In summary, we make the following contributions:

(1) The paper proposes an effective and light-weight mechanism to integrate the medical knowledge into different neural generative models, *MKA*. Besides, the specific medical knowledge graph is designed to store the medical knowledge. To the best of our knowledge, *MKA* is the first scalable work that can integrate the medical knowledge into all kinds of neural generative models, especially for large-scale pretrained model, such as *BERT-GPT*.

(2) To verify our method, we implement two models based on our mechanism, *MKA-Transformer* and *MKA-Bert-GPT.* The evaluation is carried out on 2 typical medical conversation benchmarks: $MedDialog$ [11] and $MedDG$ [12]. Our experiments show that the model combined with our method outperform previous methods in multiple automatic evaluation metrics. Besides, the *MKA-Bert-GPT* achieves the best performance on the task.

The paper will be separated into 5 parts. Section 2 Related Works will present the existed works related to medical dialogue generation tasks. Section 3 Methodology will explain the details of the proposed mechanism. Section 4 Experiment shows the experiment results and the analysis of the results. Section 5 Conclusion concludes the advantages and disadvantages of our work and its potential future works.

## 2. Related Works

Recent research on medical chatbots focus on natural language understanding which leverages different advanced natural language processing *(NLP)* techniques. In general, the medical dialogue methods can be divided into information retrieval-based methods and neural generative methods according to the types of the applied *NLP* techniques. The retrieval-based methods can be further classified into different subtypes, such as the entity inference [12][13], relation prediction [14][15], symptom matching and extraction [16][17], slot filling [18][19][20]. However, the retrieval-based methods are not so intelligent and flexible that they required a well-defined user-built question and answer (*Q&A*) pool, which can offer different potential response to different kinds of answer. In another word, the retrieval-based methods only predict the link between question to answers in the pool, instead of learning how to response to different questions like the doctors. Therefore, the neural generative methods have drawn more and more attention.

Nowadays, there is merely research on developing neural generative methods on medical domain. As an emerging research direction, most of the existed research focus on testing different neural generative models on the benchmark domain specific datasets. To well figure out the generative tasks in *NLP*, Hochreiter and Schmidhuber first proposed Long Short-Term Memory (*LSTM*) [21], which inspires multiple *LSTM-based* models [22][23][24]. Later, with the proposed of *Transformer* [25], researchers start to leverage *Transformer* units into novel dialogue generation models [26][27]. Then, a more accurate and faster mechanism *GPT* is proposed [28], different large scale dialogue generative models are developed based on it [29][30]. Meanwhile, some of the works also attempt to combine the different units to develop novel methods, where the state-of-the-art model is *BERT-GPT* model [31][32]. However, the existed the generative models for medical domain only learn the experience knowledge from the training procedure, few work effectively integrates the medical knowledge with the generative models.

## 3. Methodology

In this section, we discuss the methodology of *MKA*, which is a scalable, effective, and light-weight mechanism to integrate the medical knowledge into neural generative models, especially for large-scale pretrained model, such as *BERT-GPT*.

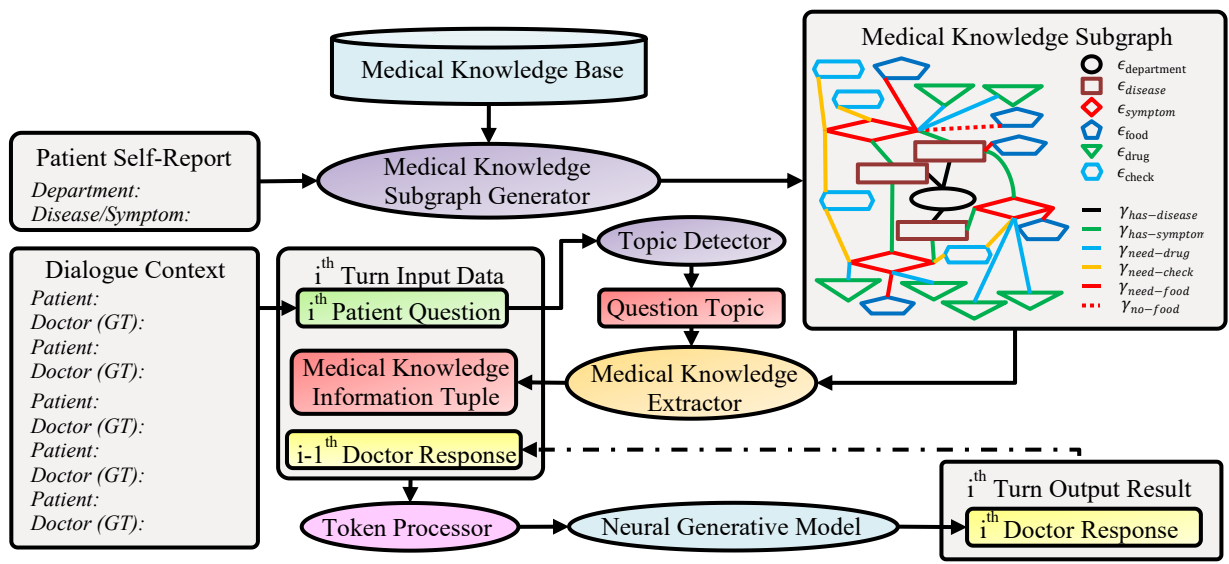

Figure 1. Framework of our scalable medical knowledge assisted generative model, *MKA*. In the figure, the ellipsoids represent the modules inside of our method. The orange ellipsoids show the automatic and scalable medical knowledge generation module. The figure legend inside of medical knowledge subgraph is corresponding to the entity types and relation types shown in Table 1 and Table 2.

As shown in figure 1, our *MKA* consists of 3 parts, including the medical knowledge generation module, token processor, and neural generative model. The medical knowledge generation module is constituted by medical knowledge subgraph generator, topic detector and medical knowledge extractor. It is aimed to generate related medical knowledge information tuple. The token processor is proposed to concatenate the medical knowledge information tuple with the dialogue context for each conversation turn. Besides, the neural generative model is leveraged for training and prediction. The details of each module will be illustrated in Sec. 3.1, Sec. 3.2, and Sec. 3.3.

*3.1 Medical Knowledge Generation Module*

The medical knowledge generation module is proposed to generate the related medical knowledge information when the doctor handles a case. Within the module, there exist three parts, including medical knowledge subgraph generator, topic detector and medical knowledge extractor. The medical knowledge subgraph generator first takes the patient self-report which contains the department and disease/symptom information described in Sec. 1 as input and generate the medical knowledge subgraph ($MKG_{sub}$) based on a global medical knowledge base ($MKG_{base}$). $MKG_{base}$ can be treated as a container which contains all the required medical professional books, while $MKG_{sub}$ stores the potential useful medical knowledge related to the specific case. Different questions will be asked in different turns for the multi-turn conversation task. To reduce the redundant information, the topic detector inputs the patient question at $i^{th}$ turn and infer what question topic it related to. With the question topic and $MKG_{sub}$, the medical knowledge extractor will extract the related medical knowledge information tuple. The details of each part will be shown as follows.

*3.1.1 Medical Knowledge Subgraph Generator*

Within the medical knowledge subgraph generator, the medical knowledge sub-base can be generated from the medical knowledge base based on the medical-related information extracted from patient self-report. In this paper, the knowledge base is represented the knowledge graph ($KG$), which is constituted by entities and relations. Besides, it is formally defined as below:

$$KG = (\mathcal{E}, \mathcal{R}, \mathcal{G} = \mathcal{E} \times \mathcal{R} \times \mathcal{E}) \quad (1)$$

where $\mathcal{E}$ represents the set of entities (e.g., persons), $\mathcal{R}$ represents the considered types of relations between entities (e.g., friendship between persons), and $\mathcal{G}$ is a set of 3-element fact tuples where each tuple represents a factual relation between two entities.

Therefore, two kinds of the medical knowledge graph ($MKG$) are proposed, including the medical knowledge base ($MKG_{base}$) generated based on [33] by removing the redundant information, and medical knowledge subgraph ($MKG_{sub}$). Both $MKG_{base}$ and $MKG_{sub}$ contain 6 types of entities and 6 types of relations as shown in Table 1 and Table 2. The entity and relation types are decided based on the working experiences of the author for the common medical conversation topics.

Table 1. Definition of entity types in medical knowledge graphs

| Entity Type | Description |
| --- | --- |
| $\mathcal{E}_{department}$ | Entities for the clinical departments |
| $\mathcal{E}_{disease}$ | Entities for the diseases |
| $\mathcal{E}_{symptom}$ | Entities for the symptoms |
| $\mathcal{E}_{food}$ | Entities for the food |
| $\epsilon_{check}$ | Entities for the drugs |
| $\epsilon_{drug}$ | Entities for the checks |

Table 2. Definition of relation types in medical knowledge graphs

| Relation Type | Description |
|---|---|
| $\gamma_{has-disease}$ | Relations between $\epsilon_{department}$ entity and $\epsilon_{disease}$ entity |
| $\gamma_{has-symptom}$ | Relations between $\epsilon_{disease}$ entity and $\epsilon_{symptom}$ entity |
| $\gamma_{need-drug}$ | Relations between $\epsilon_{disease}/\epsilon_{symptom}$ entity and $\epsilon_{drug}$ entity |
| $\gamma_{need-check}$ | Relations between $\epsilon_{disease}/\epsilon_{symptom}$ entity and $\epsilon_{check}$ entity |
| $\gamma_{need-food}$ | Relations between $\epsilon_{disease}/\epsilon_{symptom}$ entity and recommended $\epsilon_{food}$ entity |
| $\gamma_{no-food}$ | Relations between $\epsilon_{disease}/\epsilon_{symptom}$ entity and not recommended $\epsilon_{food}$ entity |

According to the definition of the entity and relation types, $MKG_{base}$ contains 26910 entities (i.e., 54 $\epsilon_{department}$, 8807 $\epsilon_{disease}$, 5998 $\epsilon_{symptom}$, 4870 $\epsilon_{food}$, 3353 $\epsilon_{check}$, and 3828 $\epsilon_{drug}$) and 158216 fact tuples regarding to different relations (i.e., 8844 $\gamma_{has-disease}$, 5998 $\gamma_{has-symptom}$, 59467 $\gamma_{need-drug}$, 39422 $\gamma_{need-check}$, 22238 $\gamma_{need-food}$, and 22247 $\gamma_{no-food}$).

As for the $MKG_{sub}$, it is specific for each case, which is generated based on Algorithm 1. Within the algorithm, two subgraphs, $G_1$ and $G_2$, are extracted from $MKG_{base}$ to constitute the $MKG_{sub}$. $G_1$ is the graph with the $\epsilon_{department}$ type entity $\epsilon_1^*$ as the root. Besides, it only contains $\gamma_{has-disease}$ and $\gamma_{has-symptom}$ two types of relations. $G_2$ is the graph with the $\epsilon_{disease}/\epsilon_{symptom}$ type entity $\epsilon_2^*$ as the root. Besides, it may contain all kinds of types of relations expect $\gamma_{has-disease}$. More details see Algorithm 1 shown in Table 3.

Meanwhile, it is worth noting that we propose a way to calculate the distance for entity matching as shown in Equation 2.

$$dist(u,v) = \begin{bmatrix} \alpha & \beta \end{bmatrix} \times \begin{bmatrix} dist_{Levenshtein}(u,v) \\ dist_{Hamming}(u,v) \end{bmatrix} \quad (2)$$

where the $\alpha$ and $\beta$ are two hyperparameters. The distance takes advantage of both the Hamming distance [34] and Levenshtein distance [35]. It can not only care about the meaning of the tokens like the Hamming distance, but also the position of the tokens like Levenshtein distance.

Table 3. Pseudo-Code of the medical knowledge subgraph generator.

| |
|---|
| **Algorithm1:** The generation of the medical knowledge subgraph |
| **Input:**<br>Medical Knowledge Base $MKG_{base} = (\mathcal{E}_b, \mathcal{R}_b, \mathcal{G}_b)$. → Eq. (1)<br>Patient Self-Report $PSR$. ($PSR \rightarrow Department$ represents the blank for the patient's ideal clinical department, and $PSR \rightarrow Disease/Symptom$ represents the blank for the description of the patient's disease or symptom.)<br>**Output:**<br>Medical Knowledge Subgraph $MKG_{sub} = (\mathcal{E}_s, \mathcal{R}_s, \mathcal{G}_s)$. → Eq. (1)<br><br>**Main:**<br>1:   **if** $PSR \rightarrow Department$ exists **then**<br>2:      Extract the $Info_1$ from $PSR \rightarrow Department$. |

3:   $\epsilon_1^* = arg \min_{\epsilon} (dist(Info_1, \epsilon))$, where $\epsilon \in \epsilon_{\text{department}}$ and $\epsilon \in \mathcal{E}_b$.    → Eq. (2)
4:   $G_1 = (\mathcal{E}_1, \mathcal{R}_1, \mathcal{G}_1)$, where $\forall \{\epsilon_{11}, \epsilon_{12}\} \subset \mathcal{E}_b$, $\forall \{g_{11}, g_{12}\} \subset \mathcal{G}_b$, $\gamma_{11} = \gamma_{has-disease}$, $\gamma_{12} = \gamma_{has-symptom}$, $\exists\ g_{11} = (\epsilon_1^*, \gamma_{11}, \epsilon_{11}) \in \mathcal{G}_1$, $g_{12} = (\epsilon_{11}, \gamma_{12}, \epsilon_{12}) \in \mathcal{G}_1$.
5:  **end if**
6:  **if** $PSR \to Disease/Symptom$ exists **then**
7:   Extract the $Info_2$ from $PSR \to Department/Symptom$
8:   $\epsilon_2^* = arg \min_{\epsilon} (dist(Info_1, \epsilon))$, where $\epsilon \in (\epsilon_{disease} | \epsilon_{symptom})$ and $\epsilon \in \mathcal{E}_b$.  → Eq. (2)
9:   $G_2 = (\mathcal{E}_2, \mathcal{R}_2, \mathcal{G}_2)$, where $\forall \{\epsilon_{21}, \epsilon_{22}, \epsilon_{23}, \epsilon_{24}\} \subset \mathcal{E}_b$, $\forall \{g_{21}, g_{22}, g_{23}, g_{24}\} \subset \mathcal{G}_b$, $\gamma_{21} = \gamma_{has-disease}$, $\gamma_{22} = \gamma_{has-symptom}$, $\forall \gamma_{23} \subset \{\gamma_{need-drug}, \gamma_{need-check}, \gamma_{need-food}, \gamma_{no-food}\}$, (1) if $\epsilon_2^* \in \epsilon_{disease}$, $\exists\ g_{22} = (\epsilon_2^*, \gamma_{22}, \epsilon_{22}) \in \mathcal{G}_2$, $g_{23} = (\epsilon_2^*, \gamma_{23}, \epsilon_{23}) \in \mathcal{G}_2$, $g_{24} = (\epsilon_{22}, \gamma_{23}, \epsilon_{24}) \in \mathcal{G}_2$; (2) if $\epsilon_2^* \in \epsilon_{symptom}$, $\exists\ g_{21} = (\epsilon_{21}, \gamma_{21}, \epsilon_2^*) \in \mathcal{G}_2$, $g_{22} = (\epsilon_{21}, \gamma_{22}, \epsilon_{22}) \in \mathcal{G}_2$, $g_{23} = (\epsilon_{21}, \gamma_{23}, \epsilon_{23}) \in \mathcal{G}_2$, $g_{24} = (\epsilon_2^*, \gamma_{23}, \epsilon_{24}) \in \mathcal{G}_2$.
10:  **end if**
11: $MKG_{sub} = (\mathcal{E}_s, \mathcal{R}_s, \mathcal{G}_s) = (\mathcal{E}_1 \cup \mathcal{E}_2, \mathcal{R}_1 \cup \mathcal{R}_2, \mathcal{G}_1 \cup \mathcal{G}_2) = G_1 \cup G_2$

### 3.1.2 Topic Detector

The medical knowledge is related to what medical topic the patient asks. As a preparation for medical knowledge extractor, the question topic should be determined first. The content in the topic set matches with the relation set (i.e., disease, symptom, drug, check, positive food, and negative food). Besides, the six key phrase sets ($KPS$) are built corresponding to six topics based on the users' experiences. It consists of some specific phrases related to the question topic. Based on it, the question detector is proposed as shown in Algorithm 2 shown in Table 4.

Table 4. Pseudo-Code of the topic detector.

**Algorithm2:** The detection of the question topic

**Input:**
Key Phrase Set $KPS = [KPS_{di}, KPS_s, KPS_{dr}, KPS_c, KPS_{rf}, KPS_{nrf}]$. ($KPS_{di}$ is the set for disease topic, $KPS_s$ is the set for symptom topic, $KPS_{dr}$ is the set for drug topic, $KPS_c$ is the set for check topic, $KPS_{rf}$ is the set for the positive food topic, $KPS_{nrf}$ is the set for the negative food topic.)
The Patient Question in $i^{th}$ Conversation Turn $PQ_i$.
Similarity Coefficient $\delta$ for checking whether the phrase is inside of the patient question

**Output:**
Question Topic Tuple in $i^{th}$ Conversation Turn $QT_i$

**Main:**
1:  $QT_i = \{\}$
2:  **for** $kps$ in $KPS$ **do**
3:   **for** $kp$ in $kps$ **do**
4:    **if** ($kp$ in $PQ_i$) | $(dist(PQ_i, kp) > \delta)$ **then**    → Eq. (2)
5:     **if** $kps = KPS_{di}$ **then**
6:      Append the 'disease' topic in $QT_i$
7:     **else if** $kps = KPS_s$ **then**
8:      Append the 'symptom' topic in $QT_i$
9:     **else if** $kps = KPS_{dr}$ **then**
10:      Append the 'drug' topic in $QT_i$
11:     **else if** $kps = KPS_c$ **then**
12:      Append the 'check' topic in $QT_i$
13:     **else if** $kps = KPS_{rf}$ **then**
14:      Append the 'recommended food' topic in $QT_i$
15:     **else if** $kps = KPS_{nrf}$ **then**
16:      Append the 'not recommended food' topic in $QT_i$

| 17: | | end if |
| 18: | end for | |
| 19: | end for | |

### 3.1.3 Medical Knowledge Extractor

The medical knowledge extractor aims to extract the related medical knowledge information tuples based on question topic and medical knowledge subgraph from the previous two parts. It extracts all entities with the specific entity type and connected with specific relation type in the subgraph. Besides, the $\epsilon_1^*, \epsilon_2^*$ extracted from patient self-report will be directly appended into the tuple, since they are also useful medical knowledge extracted from the source. The details of the algorithm are shown in Table 5.

Table 5. Pseudo-Code of the medical knowledge extractor.

**Algorithm3**: The extraction of medical knowledge information tuple

**Input:**
Medical Knowledge Base $MKG_{sub} = (\mathcal{E}_s, \mathcal{R}_s, \mathcal{G}_s)$.
Question Topic Tuple in $i^{th}$ Conversation Turn $QT_i$
Corresponding $\epsilon_{\text{department}}$ and $\epsilon_{\text{disease}}/\epsilon_{\text{symptom}}$ entities in patient self-report $\epsilon_1^*, \epsilon_2^*$   → Table. 3

**Output:**
Medical Knowledge Information Tuple in $i^{th}$ Conversation Turn $MKI_i$

**Main:**
1:   $MKI_i = \{\epsilon_1^*, \epsilon_2^*\}$
2:   **for** $qt$ in $QT_i$ **do**
3:       **if** $QT_i$ = 'disease' **then**
4:           Append all the $\epsilon_{\text{disease}}$ entities except $\epsilon_2^*$ in $MKG_{sub}$ to $MKI_i$
5:       **else if** $QT_i$ = 'symptom' **then**
6:           Append all the $\epsilon_{\text{symptom}}$ entities except $\epsilon_2^*$ in $MKG_{sub}$ to $MKI_i$
7:       **else if** $QT_i$ = 'drug' **then**
8:           Append all the $\epsilon_{\text{drug}}$ entities in $MKG_{sub}$ to $MKI_i$
9:       **else if** $QT_i$ = 'check' **then**
10:          Append all the $\epsilon_{\text{check}}$ entities in $MKG_{sub}$ to $MKI_i$
11:      **else if** $QT_i$ = 'recommended food' **then**
12:          Append all the $\epsilon_{\text{food}}$ entities connected with the $\gamma_{\text{need-food}}$ relation in $MKG_{sub}$ to $MKI_i$
13:      **else if** $QT_i$ = 'not recommended food' **then**
14:          Append all the $\epsilon_{\text{food}}$ entities connected with the $\gamma_{\text{no-food}}$ relation in $MKG_{sub}$ to $MKI_i$
15:      **end if**
16:  **end for**

### 3.2 Token Processor

Compared to general neural generative models just taking dialogue contexts as inputs, our model generates the related medical knowledge information tuple which will be also fed into the models. To achieve this goal, a token processor is proposed to reorganize the tokens based on the policy shown in Equation (2).

$$\begin{aligned} X &= MKI_i \oplus DR_{i-1} \oplus PQ_i \\ &= \left[ MKG_{sub} \text{ Unseen Features} \right] \oplus \left[ PSR \text{ Seen Features} \right] \oplus DR_{i-1} \oplus PQ_i \\ &= \left[ \text{All the elements in } MKI_i \text{ except } \varepsilon_1^* \text{ and } \varepsilon_2^* \right] \oplus \left[ \varepsilon_1^* \oplus \varepsilon_2^* \right] \oplus DR_{i-1} \oplus PQ_i \end{aligned} \quad (2)$$

where $X$ represents the sequence for neural generative models, and $PSR$ represents patient self-report. $MKI_i, DR_{i-1}, PQ_i$ represent the medical knowledge information tuple in $i^{th}$ conversation turn, the doctor response in $i - 1^{th}$ conversation turn, and the patient question in $i^{th}$ conversation turn separately. Besides, $\epsilon_1^*$ and $\epsilon_2^*$ are corresponding $\epsilon_{\text{department}}$ and $\epsilon_{\text{disease}}/\epsilon_{\text{symptom}}$ entities generated in Algorithm 1.

*3.3 Neural Generative Model*

In this paper, the neural generative model takes a source sequence $X = (x_1, x_2, x_3, ..., x_T)$ consisting of $T$ tokens generated from Section 3.2 and generate the response $Y = (y_1, y_2, y_3, ..., y_{T'})$ of length $T'$ tokens. In general, the model maximizes the generation probability of $Y$ conditioned on $X$: $p(y_1, y_2, y_3, ..., y_{T'} | x_1, x_2, x_3, ..., x_T)$ [8], and the objective function of the sequence-to-sequence generative models is defined as below. Besides, as for the multi-turn conversation tasks, the doctor response at $i^{th}$ turn will be fed into the model as the existed dialogue context for next turn.

$$p(y_1,...,y_{T'} | x_1,...,x_T) = p(y_1 | x_1,...,x_T)\prod_{t=2}^{T'} p(y_t | x_1,...,x_T, y_1,...,y_{t-1}) \qquad (3)$$

## 4. Experiments

*4.1. Experiment Settings*

Our approach is implemented in python 3.7 and pytorch 1.4.0. We implement two *MKA-Diagen* models, including *MKA-Transformer* and *MKA-BERT-GPT*. The neural generative models within them are trained with the default parameters in [11][25]. The hyperparameters $\alpha$ and $\beta$ in Equation (2) are set as 0.1 and -1, and the hyperparameters $\delta$ is set as 0.7. We perform all the experiments on the Matpool server with 11 GB NVIDIA GeForce RTX 2080 Ti. Our experiments were performed on Chinese MedDialog dataset [11] and MedDG [12] with the ratio 0.8:0.1:0.1 of training set: validation set and test set.

The *MKA-Transformer* and *MKA-BERT-GPT* compared with the baseline models (i.e., *Transformer* and *BERT-GPT*) and another typical non sequence to sequence GPT-based model [11]. We followed the automatic evaluation metrics on the datasets to evaluate the performance of our method, including perplexity, NIST-2,4 [36], BLEU-2,4 [37], METEOR [38], Entropy-4 [39], and Dist-1,2 [40]. The perplexity shows the language quality of the generated responses. NIST-n, BLEU-n, and METEOR measure the similarity between the generated responses and ground truth and Entropy-n and Dist-n measure the lexical diversity of generated responses based on n-gram matching. The model with better performance will have the lower value of perplexity, the higher value of the other metrics.

*4.2. Experiment Results and Analysis*

In this part, the experiment results are shown together with the in-depth analysis of the results. Table 6 and Table 7 show the performance on the MedDialog-CN test set and MedDG test set separately. From the tables, we make the following observations.

Table 6. Comparison of the models on MedDialog-CN, where the best results are in bold

| Model | Dialog-GPT | Transformer | BERT-GPT | MKA-Transformer | MKA-BERT-GPT |
|---|---|---|---|---|---|
| perplexity | 9.71 | 9.52 | 8.23 | 8.81 | **8.04** |
| BLEU-2 | 5.21% | 4.92% | 4.88% | 5.02% | **5.71%** |
| BLEU-4 | 1.83% | 0.90% | 0.97% | 0.99% | **1.35%** |
| NIST-2 | 0.36 | 0.42 | 0.40 | 0.43 | **0.44** |
| NIST-4 | 0.32 | 0.40 | 0.39 | 0.40 | **0.43** |
| METEOR | 12.32% | 13.11% | 12.83% | 13.4% | **13.94%** |
| Entropy-4 | 13.73 | 13.51 | 13.8 | 13.72 | **14.1** |
| Dist-1 | 0.02% | 0.03% | 0.03% | 0.03% | **0.04%** |
| Dist-2 | 2.01% | 2.02% | 2.14% | 2.11% | **2.22%** |

Table 7. Comparison of the models on MedDG, where the best results are in bold

| Model | Dialog-GPT | Transformer | BERT-GPT | MKA-Transformer | MKA-BERT-GPT |
|---|---|---|---|---|---|
| perplexity | 8.53 | 8.52 | 5.98 | 8.41 | **5.95** |
| BLEU-2 | 6.41 % | 6.30% | 7.62% | 6.62% | **8.09%** |
| BLEU-4 | 2.12% | 2.08% | 2.57% | 2.40% | **2.87%** |
| NIST-2 | 0.38 | 0.37 | 0.42 | 0.39 | **0.43** |
| NIST-4 | 0.35 | 0.35 | 0.39 | 0.38 | **0.41** |
| METEOR | 13.78% | 14.32% | 16.25% | 14.88% | **16.63%** |
| Entropy-4 | 10.56 | 10.17 | **13.38** | 10.28 | 13.37 |
| Dist-1 | 0.01% | 0.01% | **0.02%** | 0.01% | **0.02%** |
| Dist-2 | 1.72% | 1.67% | **2.00%** | 1.69% | **2.00%** |

### 4.2.1 Ablation Analysis

Table 8. Improvements of the models w. MKA compared to the baseline model on MedDialog-CN and MedDG test sets

| Dataset | Model | perplexity | BLEU-2,4 | NIST-2,4 | METEOR | Entropy-4 | Dist-1,2 |
|---|---|---|---|---|---|---|---|
| MedDialog-CN | MKA-Transformer | -0.71 | 0.10%,0.09% | 0.01, 0 | 0.29% | 0.21 | 0.00%,0.09% |
| | MKA-BERT-GPT | -0.19 | 0.83%,0.38% | 0.04, 0.04 | 1.11% | 0.3 | 0.01%,0.08% |
| MedDG | MKA-Transformer | -0.11 | 0.32%,0.32% | 0.02, 0.03 | 0.56% | 0.11 | 0.00%,0.02% |
| | MKA-BERT-GPT | -0.03 | 0.47%,0.30% | 0.01, 0.02 | 0.38% | -0.01 | 0.00%,0.00% |

Focusing on the comparison between *MKA-Transformer and Transformer* and the performance comparison between *MKA-BERT-GPT* and *BERT-GPT*, it is easy to extract Table 8. It is easy to observe that our mechanism improves the performance from all aspects on both two datasets. It means that our method is effective, and scalable to be applied to different neural generative models and different datasets.

### 4.2.2 Performance Comparison Analysis

Compared to the current state of the art models, our *MKA-BERT-GPT* outperforms all the other methods. It achieves lowest perplexity. It is because its baseline generative model, *BERT-GPT*, is pretrained on a large collection of corpora before training on the medical specific datasets. The pertaining procedure helps it to better understand the linguistic structure among words, meanwhile the medical knowledge assisted mechanism enables the model more learnable for medical conversation task. Meanwhile, as for the machine translation metrics (i.e., NIST-4, BLEU-2, BLEU-4, and METEOR), the performance of the *MKA-BERT-GPT* also is the best. It even overturns the performance comparison between BERT-GPT and Transformer. It indicates that our method highly improves the overlap between the generated response and the ground truth. Besides, although the *MKA-BERT-GPT* improves the value on diversity metrics (i.e., Entropy, Dist), the improvement is still minor. It indicates that our model cannot make a big breakthrough on the capability in generating diverse responses.

### 4.2.3 Case Study Analysis

Table 9 and Table 10 represent the generated response of the models on two examples in the MedDialog-CN and MedDG test set. Since the dataset contains some Chinese medical dialogues, the translation is provided as well as the raw contents. The response generated by *MKA-BERT-GPT* is clinically informative and accurate. It prescribes "gastrointestinal functional problem". Meanwhile it can offer the detailed suggestions with rich medical knowledge information such as what kind of vegetables and fruits is recommended. Besides, the language quality of all the models is great, since all the responses are readable. Besides, there are still some spaces for the further improvement. For example, the responses generated from the models are not that overlap with the ground truth. It is because that the ground truth is a Chinese medical response, which contains the concept of "qi", which is not that easy for a general model to understand and provide the response. However, the responses of *MKA-BERT-GPT* are still relative reasonable and also mentions the conclusion of "The blood flow is not smooth.".

Table 9. An example of generated responses on the MedDialog-CN test set

| | Patient Self-Report | 科室 Department：男科 Andrology<br>疾病/症状 Disease/Symptom：静索静脉曲张 Varicose veins |
|---|---|---|
| | Patient Question 1 | 我很怕是高位结扎手术引起的附睾炎，希望大夫能给我说说是怎么回事，该怎么处理？<br>I am very afraid that it is the epididymitis associated with the high ligation operation. I hope the doctor can tell me how to deal with it quickly. |
| Doctor Response 1 | Ground Truth | 中医药调理。<br>You can recuperate yourself with Chinese medicine. |
| | Dialog-GPT | 手术部位不会引发附睾炎。<br>The surgical part does not cause epididymitis. |
| | Transformer | 可能是近期没有充分休息，免疫力下降。建议早点休息。<br>It may be that you don't have enough rest and cause the decrease on the immunity. It is recommended to sleep early. |
| | BERT-GPT | 不要趴着睡觉，影响伤口恢复。建议多休息。<br>Do not sleep on your stomach, which will affect the recovery of the wound. It is recommended to rest more. |
| | MKA-Transformer | 可以吃营养的健康食品，比如西兰花，鸡蛋等，同时注重卫生清洁，每晚可以热敷缓解疼痛。<br>You can eat nutritious and healthy foods, such as broccoli, eggs, etc. Meanwhile, pay attention to your hygiene and cleanliness. You can heat you're the area around your wound every night to relieve yourself. |
| | MKA-BERT-GPT | 不可以吃冷饮，少吃辛辣刺激的食物，影响伤口恢复。建议多吃蔬果类，比如苹果。不要久坐不动，少穿紧身裤。<br>Do not drink cold drinks, and eat less spicy food, which will affect the recovery of the wound. It is recommended to eat more fruits and vegetables, such as apples. Don't sit for a long time and do not wear tight pants. |
| | Patient Question 2 | 请问我这是附睾炎吗？或是什么？恳请大夫您为我指点！感谢。<br>Is this epididymitis? Or what? I hope you, the doctor, to give me some advice! Thanks! |
| Doctor Response 2 | Ground Truth | 局部气血不畅。<br>The flow of the blood and qi in the local area is not that smooth. |
| | Dialog-GPT | 放宽心，只是心理作用。<br>Relax, it is just a psychological effect. |
| | Transformer | 只是术后伤口疼痛。<br>It's just that the wound is painful after the operation. |
| | BERT-GPT | 常见血流不通问题。<br>It is the common problem about the blood. |
| | MKA-Transformer | 可能是运动过量导致提睾肌过度收缩，导致疼痛。<br>Excessive exercise may cause excessive contraction of the cremaster muscle. It will make you feel painful. |
| | MKA-BERT-GPT | 血流不通，建议过几天复查伤口恢复情况，减轻心理负担。<br>The blood flow is not smooth. It is recommended to review the situation of the recovery of the wound after a few days. Besides, you need to reduce the psychological burden. |

Table 10. An example of generated responses on the MedDG test set

| | Patient Self-Report | 科室 Department：肛肠科 Anorectal Branch<br>疾病/症状 Disease/Symptom：便秘,便血 Constipation, blood in the stool |
|---|---|---|
| | Patient Question 1 | 医生你好，最近我大便总是很干燥，而且有好几次大便之后，发现有血，我这个是什么情况？<br>Hello, doctor. Recently, I have found my stools were very dry. Besides, there are blood inside of them if I poop for several times. What's wrong with me? |
| Doctor Response 1 | Ground Truth | 你好，可能存在痔疮出血！但还是建议您做个肠镜！<br>Hello, there may be bleeding from hemorrhoids! But it is still recommended that you have a colonoscopy! |
| | Dialog-GPT | 你好，这种情况容易出现内痔<br>Hello, this situation is prone to internal hemorrhoids |
| | Transformer | 你好，考虑是痔疮。<br>Hello, it is considered as hemorrhoids. |
| | BERT-GPT | 你好，可能存在痔疮，大便干结多久了？ |
| | | Hello, there may be hemorrhoids, and how long have the stools been dry and clumpy? |
| | MKA-Transformer | 你好，可能是痔疮，建议您做个肠镜。<br>Hello, it may be hemorrhoids. It is recommended that you have a colonoscopy. |
| | MKA-BERT-GPT | 你好，可能是痔疮出血。此外，建议您做个肠镜。<br>Hello, it may be bleeding from hemorrhoids. In addition, it is recommended that you have a colonoscopy. |
| Patient Question 2 | | 为什么我的大便是一粒一粒的，很干燥，像羊屎那样的？<br>Why do my stools come in pieces, very dry, like sheep feces? |
| Doctor Response 2 | Ground Truth | 大便干结是功能性问题，建议您多吃水果蔬菜，养成排便习惯！<br>The dry stool is a functional problem. It is recommended that you eat more fruits and vegetables to develop a bowel habit! |
| | Dialog-GPT | 你好，可能存在痔疮。<br>Hello, there may be hemorrhoids. |
| | Transformer | 注意饮食，多吃蔬菜水果。<br>Pay attention to your diet and eat more fruits and vegetables. |
| | BERT-GPT | 建议您多吃蔬菜水果。<br>It is recommended that you eat more fruits and vegetables |
| | MKA-Transformer | 考虑是肠胃功能性紊乱，建议您多吃水果蔬菜。<br>It is considered as a gastrointestinal dysfunction. It is recommended that you eat more fruits and vegetables, like apples. |
| | MKA-BERT-GPT | 考虑是肠胃功能问题，建议您多吃水果蔬菜，比如梨，香蕉。若还是这样的话建议您做个肠镜。<br>It is considered as the gastrointestinal functional problem. It is recommended that you eat more fruits and vegetables, such as pears and bananas. If this is still the case, it is recommended that you have a colonoscopy. |

## 5. Conclusions

In this paper, we propose a scalable Medical Knowledge Assisted mechanism, *MKA*, to assist general neural generative models, especially the large scale pretrained model, such as BERT-GPT, to achieve better performance on the medical conversation task. The mechanism introduces a medical specific knowledge graph, which contains 6 types of medical related information, including department, drug, check, symptom, disease, food. Besides, it also leverages the specific designed token concatenation policy and neural generative models. The promising experiment results have proven our mechanism is effective and scalable to different generative models on different medical conversation datasets. Besides, it also shows that *MKA-Bert-GPT* has achieved the state-of-the-art performance based on multiple automatic evaluation metrics compared to other existed models. In the future, we plan to apply the graph neural networks to extract and predict the related medical knowledge based on the medical knowledge base. Besides, it is also worthwhile to carry out the research on leverage the advantages of both information retrieve methods and the neural generative methods to build a powerful dialogue generation system.

**Funding Statement**
The authors did not receive specific funding.

**Conflicts of Interest**
The authors declare that they have no competing interest.

**Data Availability Statement**
The data used to support the findings of this study are included within the article.

## References

[1] Khan, A., Asghar, M. Z., Ahmad, H., Kundi, F. M., Ismail, S. (2017). A rule-based sentiment classification framework for health reviews on mobile social media. Journal of

Medical Imaging and Health Informatics, 7(6), 1445-1453.
[2] Deepa, N., Prabadevi, B., Maddikunta, P. K., Gadekallu, T. R., Baker, T., Khan, M. A., Tariq, U. (2021). An AI-based intelligent system for healthcare analysis using Ridge-Adaline Stochastic Gradient Descent Classifier. The Journal of Supercomputing, 77, 1998-2017.
[3] Javed, A.R., Sarwar, M.U., Beg, M.O., Asim, M., Baker, T., and Tawfik, H. (2020). A collaborative healthcare framework for shared healthcare plan with ambient intelligence. Human-centric Computing and Information Sciences, 10, 1-21.
[4] Asghar, M. Z., Khan, A., Khan, K., Ahmad, H., Khan, I. A. (2017). COGEMO: Cognitive-Based Emotion Detection from patient generated health reviews. Journal of Medical Imaging and Health Informatics, 7(6), 1436-1444.
[5] Dhanamjayulu, C., Nizhal, U. N., Maddikunta, P. K. R., Gadekallu, T. R., Iwendi, C., Wei, C., Xin, Q. (2021). Identification of malnutrition and prediction of BMI from facial images using real-time image processing and machine learning. IET Image Processing.
[6] Lee, D., Yoon, S. N. (2021). Application of Artificial Intelligence-Based Technologies in the Healthcare Industry: Opportunities and Challenges. International journal of environmental research and public health, 18: 271.
[7] Palanica A, Flaschner P, Thommandram A, Li M, Fossat Y (2019). Physicians' Perceptions of Chatbots in Health Care: Cross-Sectional Web-Based Survey. J. Med. Internet Res. 2019, 21: 1-10.
[8] Habib, Fakih Awab, Shakil, Ghare Shifa, Iqbal, Shaikh Sabreen Mohd, Sajid, Shaikh Tasmia Abdul (2021) Survey on Medical Self-Diagnosis Chatbot for Accurate Analysis Using Artificial Intelligence. In: Proceedings of Second International Conference on Smart Energy and Communication. Singapore. pp: 587-593.
[9] Mohiyuddin, A., Javed, A.R., Chakraborty, C., Muhammad Rizwan, Maryam Shabbir and Jamel Nebhen (2021). Secure Cloud Storage for Medical IoT Data using Adaptive Neuro-Fuzzy Inference System. Int. J. Fuzzy Syst, 5352108.
[10] Chen, H., Liu, X., Yin, D., Tang, J. (2017). A Survey on Dialogue Systems: Recent Advances and New Frontiers. In: Proceedings of Second International Conference on Smart Energy and Communication. In: SIGKDD. New York. pp: 1931-0145.
[11] Chen, S., Ju, Z., Dong, X., Fang, H., Wang, S., Yang, Y., Zeng, J., Zhang, R., Zhang, R., Zhou, M., Zhu, P., Xie, P. (2020). MedDialog: A Large-scale Medical Dialogue Dataset. In: Proceedings of the 2020 Conference on Empirical Methods in Natural Language Processing (EMNLP). Online: pp: 9241-9250.
[12] Guangtao Zeng, Wenmian Yang, Zeqian Ju, Yue Yang, Sicheng Wang, Ruisi Zhang, Meng Zhou, Jiaqi Zeng, Xiangyu Dong, Ruoyu Zhang, Hongchao Fang, Penghui Zhu, Shu Chen and Pengtao Xie (2020). MedDG: A Large-scale Medical Consultation Dataset for Building Medical Dialogue System. In: Proceedings of the 2020 Conference on Empirical Methods in Natural Language Processing. Online. pp: 9241-9250.
[13] Du, N., Chen, K., Kannan, A., Tran, L., Chen, Y. and Shafran, I. (2019) Extracting Symptoms and their Status from Clinical Conversations. In Proceedings of the 57th Annual Meeting of the Association for Computational Linguistics. Florence. pp:915–925
[14] Lin, Xinzhu, He, Xiahui, Chen, Qin, Tou, Huaixiao, Wei, Zhongyu and Chen, Ting (2019). Enhancing Dialogue Symptom Diagnosis with Global Attention and Symptom Graph. In: Proceedings of the 2019 Conference on Empirical Methods in Natural

Language Processing and the 9th International Joint Conference on Natural Language Processing (EMNLP-IJCNLP). Hong Kong. pp:5033-5042.
[15] Du, N., Wang, M., Tran, L., Lee, G., and Shafran, I. (2019). Learning to Infer Entities, Properties and their Relations from Clinical Conversations. In: Proceedings of the 2019 Conference on Empirical Methods in Natural Language Processing and the 9th International Joint Conference on Natural Language Processing (EMNLP-IJCNLP). Hong Kong. pp: 4978–4989.
[16] A. Sarker, A. Z. Klein, J. Mee, P. Harik, and G. Gonzalez Hernandez (2019). An interpretable natural language processing system for written medical examination assessment. Journal of Biomedical Informatics, vol. 98. pp: 103268.
[17] Xu, L., Zhou, Q., Gong, K., Liang, X., Tang, J., Lin, L. (2019). End-to-End Knowledge-Routed Relational Dialogue System for Automatic Diagnosis. In: The Thirty-Third AAAI Conference on Artificial Intelligence (AAAI-19). Hawaii. pp: 7346-7353.
[18] Shi, X., Hu, H., Che, W., Sun, Z., Liu, T., Huang, J. (2020). Understanding Medical Conversations with Scattered Keyword Attention and Weak Supervision from Responses. In: AAAI. New York. pp: 8838-8845.
[19] Liao, K., Liu, Q., Wei, Z., Peng, B., Chen, Q., Sun, W., Huang, X. (2020). Task-oriented Dialogue System for Automatic Disease Diagnosis via Hierarchical Reinforcement Learning. https://arxiv.org/abs/2004.14254.
[20] Wei, Z., Liu, Q., Peng, B., Tou, H., Chen, T., Huang, X., Wong, K., Dai, X. (2018). Task-oriented Dialogue System for Automatic Diagnosis. In: Proceedings of the 56th Annual Meeting of the Association for Computational Linguistics (Short Papers). Melbourne. pp: 201-207.
[21] Sepp Hochreiter, Jürgen Schmidhuber (1997). Long Short-Term Memory. Neural Computation, 8: 1735-1780.
[22] Ilya Sutskever, Oriol Vinyals, Quoc V Le (2014). Sequence to sequence learning with neural networks. In: Neural Information Processing Systems (NIPS). Montreal. pp: 3104–3112.
[23] Williams, J., Zweig, G. (2016). End-to-end LSTM-based dialog control optimized with supervised and reinforcement learning. https://arxiv.org/abs/1606.01269.
[24] Alex Sherstinsky (2020). Fundamentals of Recurrent Neural Network (RNN) and Long Short-Term Memory (LSTM) Network. Physica D: Nonlinear Phenomena, 404: 132306.
[25] Ashish Vaswani, Noam Shazeer, Niki Parmar, Jakob Uszkoreit, Llion Jones, Aidan N Gomez, Łukasz Kaiser, Illia Polosukhin (2017). Attention is all you need. In: Neural Information Processing Systems (NIPS). Long Beach. pp: 5998–6008
[26] Xiangyu Zhao, Longbiao Wang, Ruifang He, Ting Yang, Jinxin Chang, Ruifang Wang (2020). Multiple Knowledge Syncretic Transformer for Natural Dialogue Generation. In: Proceedings of The Web Conference 2020 (WWW '20). New York. pp: 752–762.
[27] Dongdong Li, Zhaochun Ren, Pengjie Ren, Zhumin Chen, Miao Fan, Jun Ma, Maarten de Rijke (2021). Semi-Supervised Variational Reasoning for Medical Dialogue Generation. In: Proceedings of the 44th International ACM SIGIR Conference on Research and Development in Information Retrieval (SIGIR '21). New York. pp: 544–554.
[28] Radford, A., Narasimhan, K. (2018). Improving Language Understanding by Generative Pre-Training. https://s3-us-west-2.amazonaws.com/openai-assets/research-covers/language-unsupervised/language_understanding_paper.pdf.

[29] Zhang, Y., Sun, S., Galley, M., Chen, Y., Brockett, C., Gao, X., Gao, J., Liu, J., Dolan, W. (2020). DIALOGPT: Large-Scale Generative Pre-training for Conversational Response Generation. In: ACL. Online. pp: 102-112.
[30] Devlin, J., Chang, M., Lee, K., Toutanova, K. (2019). BERT: Pre-training of Deep Bidirectional Transformers for Language Understanding. In: Proceedings of NAACL-HLT 2019. Minneapolis. pp: 4171-4186.
[31] Wu, Q., Li, L., Zhou, H., Zeng, Y., Yu, Z. (2020). Importance-Aware Learning for Neural Headline Editing. In: AAAI 2020. New York. pp: 9282-9289.
[32] Lewis, M., Liu, Y., Goyal, N., Ghazvininejad, M., Mohamed, A., Levy, O., Stoyanov, V., Zettlemoyer, L. (2020). BART: Denoising Sequence-to-Sequence Pre-training for Natural Language Generation, Translation, and Comprehension. In: ACL 2020. Seattle. pp: 7871-7880.
[33] H.Y. Liu (2017). QA Based on Medical Knowledge Graph. https://github.com/liuhuanyong/Q ASystemOnMedicalKG.
[34] Norouzi, M., Fleet, D.J., Salakhutdinov, R. (2012). Hamming Distance Metric Learning. In: Advances in Neural Information Processing Systems. Lake Tahoe. pp: 1061–1069.
[35] Levenshtein, V. (1965). Binary codes capable of correcting deletions, insertions, and reversals. Soviet physics, 10: 707-710.
[36] Doddington, G. (2002). Automatic evaluation of machine translation quality using n-gram co-occurrence statistics. In: Proceedings of the second international conference on Human Language Technology Research. Edmonton. pp: 71-78.
[37] Papineni, K., Roukos, S., Ward, T., Zhu, W. (2002). Bleu: a Method for Automatic Evaluation of Machine Translation. In: ACL. Philadelphia. pp: 311-318.
[38] Lavie, A., Agarwal, A. (2007). METEOR: An Automatic Metric for MT Evaluation with High Levels of Correlation with Human Judgments. WMT ACL. Prague. pp: 228–231
[39] Zhang, Y., Galley, M., Gao, J., Gan, Z., Li, X., Brockett, C., Dolan, W. (2018). Generating Informative and Diverse Conversational Responses via Adversarial Information Maximization. In: NeurIPS. Montreal. pp: 1815–1825.
[40] Li, J., Galley, M., Brockett, C., Gao, J., Dolan, W. (2016). A Diversity-Promoting Objective Function for Neural Conversation Models. In: NAACL. San Diego. pp: 110–119.